\documentclass[letterpaper, 10 pt, conference]{ieeeconf}  

\IEEEoverridecommandlockouts                              

\usepackage[usenames, dvipsnames]{color}
\usepackage{graphicx}
\usepackage{amsmath}
\usepackage{amssymb}
\usepackage{todonotes}
\usepackage{verbatim}
\usepackage{subcaption}
\usepackage[ruled]{algorithm2e}
\usepackage{url}
\usepackage[normalem]{ulem}
\usepackage{multirow}

\newcommand{\etal}{\textit{et al}.}

\newcommand{\ie}{\textit{i}.\textit{e}. }

\newcommand{\norm}[1]{\left\lVert#1\right\rVert}

\title{\LARGE \bf
Learning Movement Assessment Primitives\\ for Force Interaction Skills
}

\author{Xiang Zhang, Athanasios S. Polydoros and Justus Piater
	\thanks{All the authors are with Intelligent \& Interactive Systems (IIS), Department of Computer Science, Universit\"at Innsbruck, 6020 Innsbruck, Austria
		{\tt\footnotesize \{firstname.lastname\}@uibk.ac.at}}%
\thanks{The research leading to these results is funded by the Austrian Federal Ministry of Transport,  Innovation and Technology (BMVIT) under the program "ICT of the Future" and project number 855425, FlexRoP.}
}

\begin{document}

\maketitle
\thispagestyle{empty}
\pagestyle{empty}

\begin{abstract}

We present a novel, reusable and task-agnostic primitive for assessing the outcome of a force-interaction robotic skill, useful e.g.\ for applications such as  quality control in industrial manufacturing. The proposed method is easily programmed by kinesthetic teaching, and the desired adaptability and reusability are achieved by machine learning models. The primitive records sensory data during both demonstrations and reproductions of a movement. Recordings include the end-effector's Cartesian pose and exerted wrench at each time step. The collected data are then used to train Gaussian Processes which create models of the wrench as a function of the robot's pose. The similarity between the wrench models of the demonstration  and the movement's reproduction is derived by measuring their Hellinger distance. This comparison creates features that are fed as inputs to a Naive Bayes classifier which estimates the movement's probability of success. The evaluation is performed on two diverse robotic assembly tasks -- snap-fitting and screwing --  with a total of 5 use cases, 11 demonstrations, and more than 200 movement executions. The performance metrics prove the proposed method's capability of generalization to different demonstrations and movements.



\end{abstract}

\section{INTRODUCTION}

Despite the large research interest on generic motion representation 
models in robotics  (e.g. motion 
primitives)\cite{ijspeert2013dynamical}, the development of 
corresponding methods for 
assessing the robot-environment interaction is rather poor. The 
significance of such models rises by the need of robotic manufacturing 
processes to rapidly evaluate the effect of motions. Especially, mass 
customized production demands smart, adaptable and collaborative robotic 
systems instead of hard-coded caged robots. Therefore, such robotic 
systems have to self-evaluate their performance via methods which can 
adapt to a 
large variety of tasks by minimum reprogramming. Those needs can be met by 
developing robots equipped with abstract and easily adaptable 
functionalities: the primitives.

 Primitives are commonly trained by imitation learning or 
 Programming by Demonstration (PbD) \cite{billard2008robot} 
 methods which facilitates fast re-programming by non-experts.
In collaborative robots, a trajectory can be assigned much more 
intuitively by kinesthetic 
teaching \cite{zimmermann2016industrial} compared to hard-coded 
programming. The data collected during kinesthetic teaching are used to 
adapt 
the primitives on the desired task via supervised or reinforcement 
learning 
\cite{billard2016robot}. Thus, kinesthetic teaching, in conjunction with 
machine 
learning approaches, provides an appealing method for non-experts to 
program a robotic system.

Nevertheless, solely the motion primitives are not sufficient to guarantee successful task 
executions since uncertainties from the environment may cause failures. Therefore, an additional model is required to assess the outcome which can enable the robotic system 
to detect its failures and 
even self-improve its skills via reinforcement learning. Thus, a 
skill-based
learning system should include both assessment and motion primitives in order to increase its adaptability and performance.

\begin{figure}
	\centering
	\includegraphics[width=0.38\paperwidth]{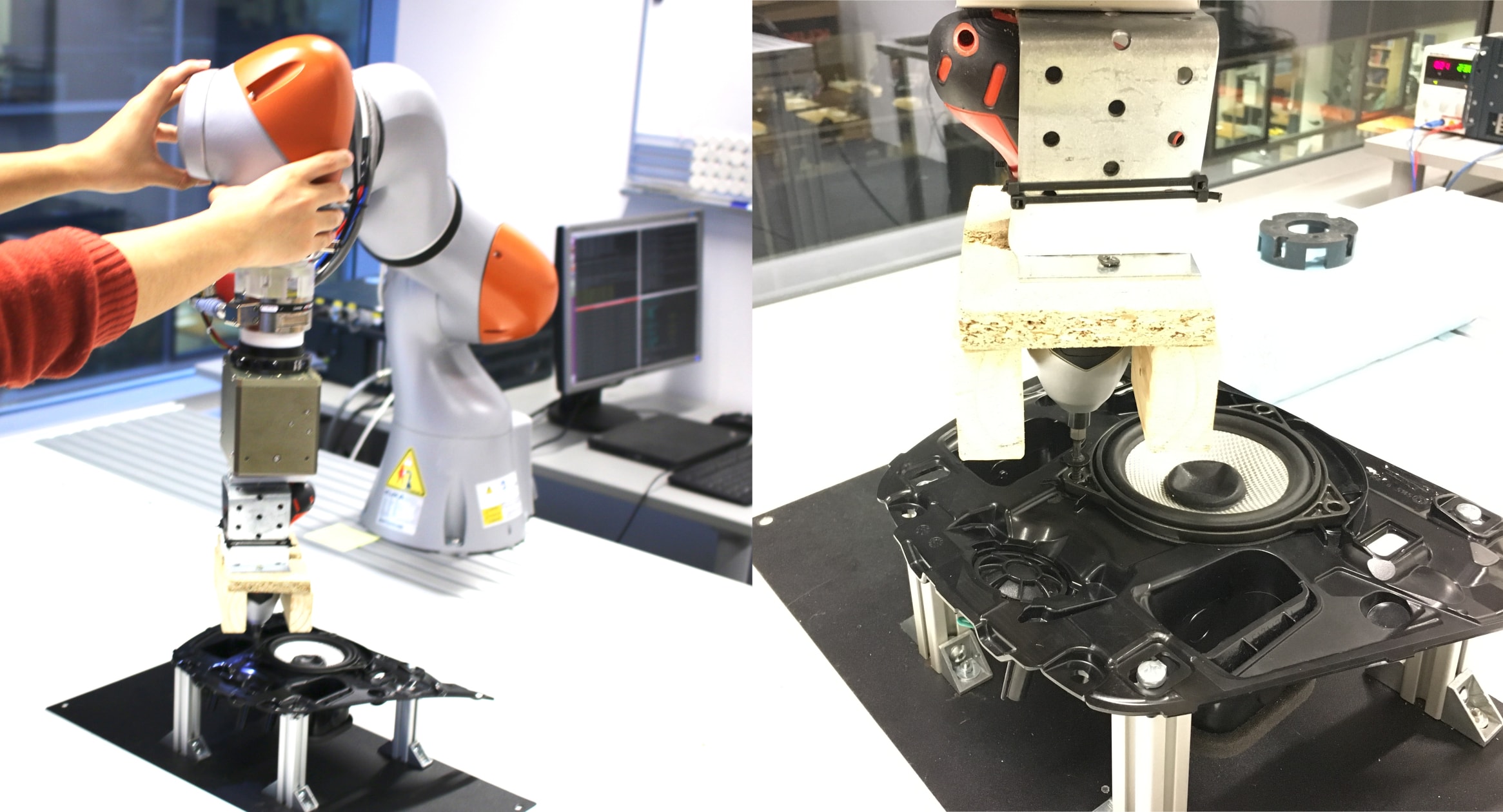}
	\caption{A KUKA iiwa equipped with an ATI F/T sensor learns how to perform a task from human demonstrations (left), and reproduce the task in novel situation (right).  Motion Assessment Primitives (MAPs) are employed for assessing the movement's reproduction performance.}
	\label{fig::intro}
\end{figure}

Therefore, this paper introduces the concept of Movement Assessment Primitives 
(MAPs). Similar to dynamic movement primitives (DMPs), MAPs are abstract elementary blocks 
which are automatically adapted on various tasks via machine learning in 
order to evaluate the motions of robotic manipulators. Furthermore, they are generic enough to be applied on a wide variety of tasks without the 
need of reprogramming. Thus, they significantly differ from
traditional methods of motion assessment in skill-based systems which require knowledge about the performed task \cite{rovida2017skiros}.

In this paper, MAPs are kinesthetically taught from demonstrations and capture the 
pose 
and applied wrench on the end-effector during the robot's movement. The aforementioned data are used to create wrench 
models which 
represent the expected wrench at each pose of the manipulator. The 
wrench models, which derive from the expert demonstration, are compared with those created from a movement 
reproduction. Their similarity measurement is treated as a feature of a binary probabilistic 
classifier which assigns a probability of success for each reproduced movement.
The proposed method is evaluated on five use-cases of robotic assembly 
operations, namely, on three mechanisms of snap-fitting and two screwing.

Thus, the main \emph{contribution} of this paper is a 
novel task-agnostic machine learning approach for automatic assessment 
of 
force-interaction skills. The proposed method focuses on the effect of 
the robot-environment interaction, which is the observed wrench, 
rather than the cause, \ie the trajectory itself. Furthermore, there is 
no 
assumption regarding the performed task and hence it can be part of any 
force-interaction skill.  Moreover, it is able to  generalize on a 
variety 
of movements with different demonstrations, initial 
and target states.

\section{Related work} \label{sec::related}

The assessment of a task execution requires modeling of sensory 
data. Such data-driven approaches are followed for fault-detection at 
process monitoring applications. In this field, the system is modeled by 
Ordinary Differential Equation (ODE)  
\cite{xie2015fault}. However, in PbD, velocity and acceleration vary 
for different human 
demonstrations and also a human is able to apply arbitrary force, but 
the force applied by a robot is 
limited by constrains such as motors, safe and impedance settings 
\cite{montebelli2015handing} etc. Therefore,
the robot cannot identically reproduce the force demonstrated by a human 
which makes the modeling of demonstrations via ODE rather challenging.

An alternative solution to assess the task outcome is statistically 
setting a threshold regarding the wrench signature and 
the pose trajectories. Costa \etal \cite{costa2015online} verify the success of a process by setting a threshold for eccentricity and 
typicality, \ie  distance metrics for time series. Haidu \etal 
\cite{haidu2015learning} define lower and upper bounds of the trajectory 
profile based on successful trails. Thus, trajectory profiles that 
exceed the threshold indicates a failure. Nevertheless, the movement 
reproduction varies for different task parameters (e.g. start and goal 
state) and demonstrations. In those cases, the bounds have to be changed 
accordingly which requires reprogramming.

Rojas \etal \cite{rojas2013towards} identify key segments of the wrench 
signals and create a task-specific hierarchical taxonomy 
based on its time derivative. The task outcome is indicated by the 
states at the highest level of this taxonomy. 
However, the segmentation threshold is predefined, and the 
taxonomy associated with segments is manually created which 
significantly reduces its applicability to different tasks.

Haidu \etal \cite{haidu2014learning} segment the task in smaller pose 
trajectories and train a Hidden Markov Model (HMM) 
using each segment as a state. They identify promising states  
by aligning the segmented samples to the HMM, 
which contains key information for failure detection. A predictor is 
trained to assess the task from the aligned data at those promising states.
Similarly, Di Lello \etal \cite{di2013bayesian} segment the wrench 
signatures by the control strategy of a finite state machine.
Each wrench signal is treated as a Bayesian time-series model. A HMM is 
trained via a Bayesian non-parametric method to detect the deviation 
from the successful execution. Both of those two methods require the predefined 
segmentation of the trajectory. However,
there is often no such clear segmentation such as Zero Velociy Crossing point 
\cite{fod2002automated} or contact events that are plausible in human 
demonstrations.

In the field of imitation learning, Calinon \etal 
\cite{calinon2007learning} evaluate a movement in order to optimize a controller. Firstly they align the recorded demonstrations with 
Hidden Markov Model (HMM) and project them 
into a latent space via PCA. Then they measure the weights of each 
vector in the latent space by the variations in 
multiple demonstrations subject to the same task constraint. Eventually, 
a similarity measure is defined by the weighted 
sum of Euclidean distances between a new trajectory and the successful 
one. The application of this 
method does not involve wrench in the task and also
requires a preprocessing procedure which increases the complexity of 
implementation.

In the field of reinforcement learning, Pastor \etal \cite{Pastor2011} 
propose an algorithm which 
learns the outcome of an assessment system given a predefined threshold 
based on a reward function. The system is merged with reinforcement 
learning framework. However, the definition of the reward function may 
be challenging for certain tasks and also limits the method's re-usability.

The aforementioned methods assess the task outcome with regard to wrench signatures 
or pose profiles separately. Thus, they do not take into account any 
dependence between wrench and state which characterizes a 
force-interaction task. 
Contrary, the proposed method takes into account this dependency by 
creating a wrench model which maps state to wrenches. 
Furthermore, the relative state w.r.t the motion's goal state is used as 
input to the proposed model. In such way is achieved the desired 
generalization of the method to different starting and goal states. 
Moreover, a similarity score between the wrench models of the 
movement demonstration and reproductions is used in order to assess the 
movement. This facilitates generalization across multiple 
demonstrations of the same task, since the wrench 
models of a specific task should not significantly vary between demonstrations. Furthermore, it is purely data-driven and thus there is no assumption regrading the executed task. To the best of our knowledge, the proposed method is the only assessment approach which is able to generalize on a variety of movement reproductions, demonstrations and also it can be applied to different tasks without reprogramming.

\section{Motion Assessment Primitives }

\begin{figure}
\centering
\includegraphics[width=0.35\paperwidth]{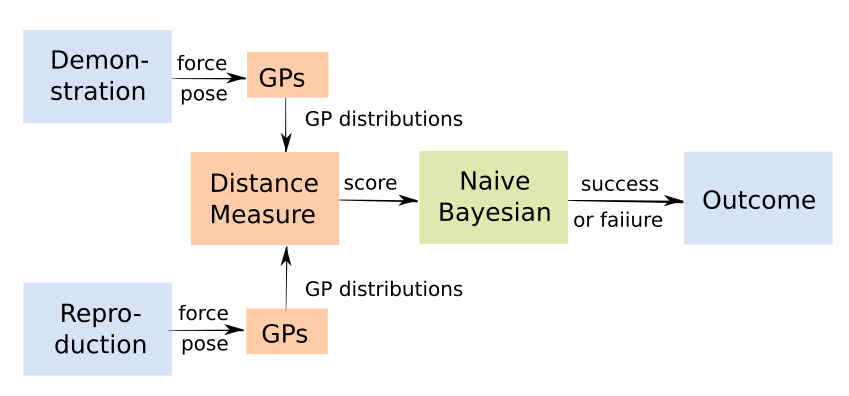}
\caption{The overall scheme of the movement assessment primitives. We train Gaussian Processes (GPs) models from  a movement  demonstration and a set of reproductions. Then their distances are calculated and used as input features for training a probabilistic classifier.}
\label{fig::structure}
\end{figure}

We formulate the wrench models as Gaussian Process (GP) which learn a mapping from the pose at each time-step to each dimension of the observed wrench. Then we extract features for a given task by measuring similarities between the GP distributions 
of a movement reproduction and its corresponding demonstration via the Hellinger distance metric. Consequently, the extracted feature vectors are used to train a  Naive Bayes 
classifier which creates the assessment model. The structure of the proposed method is illustrated in Fig. \ref{fig::structure}.

\subsection{GPs as Wrench Models}

Training of MAPs requires data about the pose, time and end-effector 
wrench during a movement demonstration and its reproductions. For 
each recorded demonstration with time stamps 
$\pmb t \in \mathbb{R}^{N}$, we assume that its corresponding reproduction 
have the same duration with $N$ number of time-steps. This assumption can be straightforwardly fulfilled 
by applying data processing methods such as 
Dynamic Time Warping \cite{gienger2010imitating}. 

The end-effector's pose at each time 
step is calculated with reference to the final desired pose. This happens to make the method invariant to
different starting and goal states. Thus matrices  $\pmb X \in \mathbb{R}^{N \times 3}$ and $\pmb Q \in 
\mathbb{R}^{N \times 4} $ represent the relative Cartesian position and orientation w.r.t the target pose at each time-step $ N $ of the movement. The inputs' matrix of the wrench models are
represented as: $ \pmb D = \left[\pmb t,\ \pmb X,\ \pmb Q\right]$

The model's output is the applied wrench, thus the target values consist of the observed 
wrench signals $\pmb W \in \mathbb{R}^{N \times 6}$ during the movement.
Since a wrench signal is a six dimensional vector each dimension $ 
w_k$ is modeled separately as: 
\begin{equation}
w_k = f_k \left(\pmb D\right) + \varepsilon
\label{eq:GPmodel}
\end{equation} where $k$ denotes the index of wrench dimensions and $ \varepsilon $ noise of zero mean and $ \sigma^2 $ variance.
 
This mapping is modeled by a Gaussian Process with zero mean and a covariance function:
\begin{equation}
f_k(\mathcal{D}) \thicksim  \mathcal{GP}(\pmb 0,\pmb K_k(\pmb D))
\label{eq:contact_model}
\end{equation}
where $\pmb K_k(\pmb D)$ is the co-variance matrix of each wrench component with elements:
\begin{equation}
\pmb K_k = \text{k}_{n, m}(\pmb d_n, \pmb d_m)
\end{equation}
where $\pmb d_n$ is the $n$th row of data matrix $\pmb D$ and $\text{k}_{n, m}$ is the $(n,m)$ element  of $ \pmb K $.

The covariance matrix specifies the properties of the distribution 
from which the function $f_k(\pmb D)$ is sampled.  
We use a square-exponential kernel
function and a constant noise as follows:
\begin{align*}
 &\text{k}(\pmb d_n, \pmb d_m) = \\
 &\theta_0 \exp\left( -\frac{\theta_1}{2}(\norm{\pmb t_n - \pmb t_m}^2+\norm{\pmb x_n - \pmb x_m}^2 +\phi_e) \right) + \sigma^2_{m,n}
\end{align*}
where $\sigma_{m,n} = 0$, for $m \neq n$ specifies the noise variance of the data. 
The variable $\phi_e$ is the angular distance between the quaternion components of $ \pmb d_n $ and $ \pmb d_m $.
Hyper-parameters $\sigma^2$, $\theta_0$ and $\theta_1$ are components of 
vector $\pmb \theta$ which are optimized to fit the data by maximizing 
the log marginal likelihood:
\begin{equation}
\ln p(\pmb w_k | \pmb \theta, \pmb{D}) = -\frac{1}{2} \ln |\pmb K| - \frac{1}{2} \pmb w_k^\text{T} \pmb K^{-1} \pmb w_k - \frac{N}{2}\ln (2\pi).
\label{eq::likelihood}
\end{equation}
where $ \pmb w_k $ is 
a column vector which contains the observations of the $ k^\text{th} $ 
wrench component.  As result, we obtain the optimized kernel matrix and thus the GP distribution over the latent function of wrench model (\ref{eq:contact_model}).

\begin{figure}

\includegraphics[width=0.4\paperwidth]{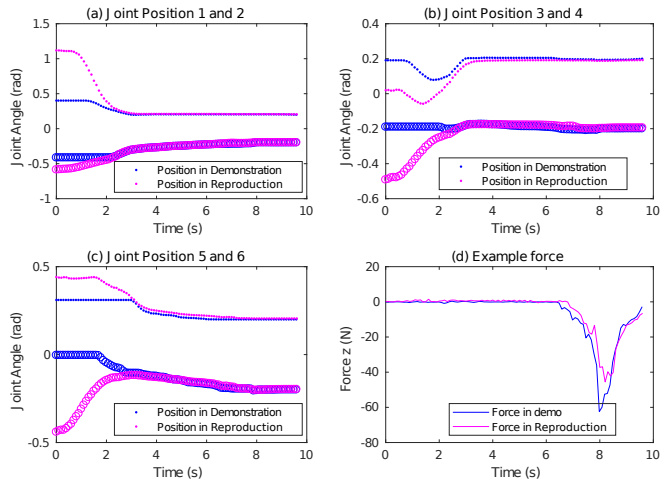}
\caption{Example illustrating the properties of GP models. (a--c) 6 pairs of position signals of a
 demonstration (blue) and corresponding reproduction (magenta) that initially differ greatly before converging.
 (d) A pair of slightly-different force signals to be modeled by MAPs.}
\label{fig::likihood}
\end{figure}

The GP models are invariant to different starting states of a movement. This fact is elaborated schematically in Fig.\ref{fig::likihood} where 6 position profiles of a movement demonstration (magenta) and a reproduction (blue) are illustrated ( Fig.\ref{fig::likihood} (a) (b) (c)). The movements start at different states compared to the demonstration but they quickly converge. Meanwhile, the exerted force profile in Fig.\ref{fig::likihood} (d) is zero at the beginning of the movement and takes values only near the movement's goal state. Thus, GPs model zero wrench at the beginning of movements regardless of their initial state. Furthermore, the use of a zero-mean GP (\ref{eq:contact_model}) facilitates this
generalization. The model assumes \textit{a priori} that states of the movement reproduction which are distant to the demonstrated states do not provide any wrench. 

Furthermore, the optimization of the log likelihood \eqref{eq::likelihood} depends on  the first two terms since the last one is constant. The first penalizes complex models while the second rewards the fit to the training data. Thus, the variation of position signals does not change the second term of the likelihood when 
the exerted wrench is zero (starting states of the movement).
Consequently, the parameters will not be affected during optimization. Furthermore,
the influence of various start positions on the first term in 
\eqref{eq::likelihood} is also limited since this term penalizes the model's complexity. As a result, the model is determined 
by the force and position signals during the contact between the manipulator and the environment.

\subsection{Similarity Features}

The assessment primitives should be applicable to movement reproductions of any demonstration. Also, a successful movement reproduction should
exert similar wrench with the demonstration. Nevertheless, not all the wrench components are significant for assessing a movement. Therefore, a separate similarity measurement is calculated for each wrench component and is used as feature representing the task's characteristics.
Thus, the similarity measurements derive based on the previously introduced wrench models of a reproduction and the corresponding demonstration.

Among several options, we compare the GP models
between demonstration and movement reproductions with Hellinger distance 
\cite{lourenzutti2014hellinger} which is a symmetric 
measurement of probability distributions' 
bounded between 0 and 1. Since the GP has analytic 
integral and we define a zero mean for the model, the 
Hellinger distance of two GPs is  written as:
\begin{equation}
 h_k\left(\mathcal{GP}_k^\text{demo},\mathcal{GP}_k^\text{rep}\right) = \sqrt{1 - \frac{\sqrt[4]{|\pmb K_k^\text{demo}| |\pmb K_k^\text{rep}|} }{\sqrt{\frac{1}{2}|\pmb K_k^\text{demo}+\pmb K_k^\text{rep}|}}}
\end{equation}
where $\mathcal{GP}_k^\text{demo}$ and $\mathcal{GP}_k^\text{rep}$ are 
the GPs' distributions of $k^\text{th}$ wrench component for the 
movement's demonstration and reproduction respectively.

However, different demonstrations of the same task may produce various amount of wrench. In order to cope with that, the Hellinger distance of each wrench component is weighted by the total wrench dissimilarity as: 
\begin{equation}
m_k =  \frac{h_k\left(\mathcal{GP}_k^\text{demo},\mathcal{GP}_k^\text{rep}\right)}{\sum_{k=1}^6 h_k\left(\mathcal{GP}_k^\text{demo},\mathcal{GP}_k^\text{rep}\right)}.
\label{eq::norm}
\end{equation}

Thus, each movement reproduction is represented by a six-dimensional features' vector $ \pmb m $ which describes the rate of dissimilarity for each wrench component. This results to a low dimensional feature vector which represents both the high-dimensional demonstrated and reproduced movements.

\subsection{Assessment Classifier}

A set of labeled similarity features $ \pmb m_l $ are used in order 
to train the assessment classifier, where the label $ l $ denotes either 
success or failure. Thus, a Na\"ive Bayes classifier is  trained with both successful and failed examples where each component of the feature vector is modeled by a normal distribution as:
\begin{equation}
p(m_k|c_l) = \mathcal{N}(m_k|\hat{\mu}_{kl},\hat{\sigma}_{kl}).
\label{eq::NBG}
\end{equation}
 given its class label $ c_l $. The parameters $ \hat{\mu}_{kl},\ \hat{\sigma}_{kl} $ of \eqref{eq::NBG} derive by maximum likelihood of estimation.

Given a new movement descriptor $ \pmb m^* $, the probability to belong in class $ c_l $ is calculated by :
\begin{equation}
p(c_l|\pmb m^*) \propto p(c_l)\prod_{k=1}^{6} p(m_k^*|c_l)
\label{eq::nb}
\end{equation}
where $p(c_l)$ is the prior probability of the class and the $p(c_l|\pmb m^*)$ indicates the likelihood of feature $ \pmb m^* $ to belong in class $ c_l $

The movement reproduction cannot be exactly identical to human demonstration due to physical limitations. For example,
iiwa robot has an impedance setting in the force mode for safety reasons, which bounds the commanded force. Such physical limitations in the robot will result to a constant similarity among movement reproductions, which does not affect the classifier.
Additionally, Naive Bayes also aggregates the similarities from different wrench dimensions since, according to \eqref{eq::nb}, the distributions of dimensions with less variance have bigger impact on the classification.

The training and application of our method for motion assessment is shown in Algorithm \ref{agr:CUSUM}.
\begin{algorithm}

        \SetKwInOut{Input}{Given}
    \SetKwInOut{Output}{Output}
    \underline{Train Movement Assessment Primitives}\;
     \Input{ $\pmb D^\text{demo},\pmb W^\text{demo}$ demonstrated movement data,
     \\~$\pmb D^\text{rep},\pmb W^\text{rep}$ movement reproductions' data, \\ ~$\pmb l$ class labels of reproductions. }
    \Output{Naive Bayes model}
    \begin{itemize}
    	\item Subtract the movement's goal pose from $\pmb D^\text{demo}$
    	\item Create $ k $ $\mathcal{GP}^\text{demo}$ models for each wrench components
    	\item \textbf{for} each movement reproduction
    		\begin{itemize}
    			\item  Subtract the movement's goal pose from $\pmb D^\text{rep}$
    			\item Create $ k $ $\mathcal{GP}^\text{rep}$ models of each wrench component.
    			\item Obtain similarity features  $ m_k $ through Hellinger Distances between
		$\mathcal{GP}_k^\text{rep}$ and $\mathcal{GP}_k^\text{demo}$
    		\end{itemize}
    		\textbf{end}
    	\item Train a Na\"ive Baysian classifier given similarity features $ \pmb m $ and class labels $ \pmb l$
    \end{itemize}
 
  \underline{Apply Movement Assessment Primitives}\;
  \Input{  $\pmb D^\text{demo},\pmb W^\text{demo}$ demonstrated movement data,\\
     ~$\pmb D^*,\pmb W^*$ data of movement to be assessed}
    \begin{itemize}
    	\item Subtract the movement's goal pose from $\pmb D^{(demo)}$
    	\item Subtract the desired goal pose from $\pmb D^*$\;
    	\item Create $ k $ $\mathcal{GP}^\text{demo}$ models of each wrench component.\;
    	\item  Create $ k $ $\mathcal{GP}^\text{*}$ models of each wrench component\;
    	\item  Obtain similarity features  $ m_k^\text{*} $ through Hellinger Distances between
    	$\mathcal{GP}_k^\text{*}$ and $\mathcal{GP}_k^\text{demo}$\;
		\item Obtain class label $l$ based on $\pmb m^*$ via the Na\"ive Bayes classifier
    \end{itemize}
  \caption{Motion Assessment Primitives}
   \label{agr:CUSUM}
 \end{algorithm}
\vspace{-10pt}

\section{Evaluation}
\begin{figure}[t]
\centering
\includegraphics[width=0.38\paperwidth]{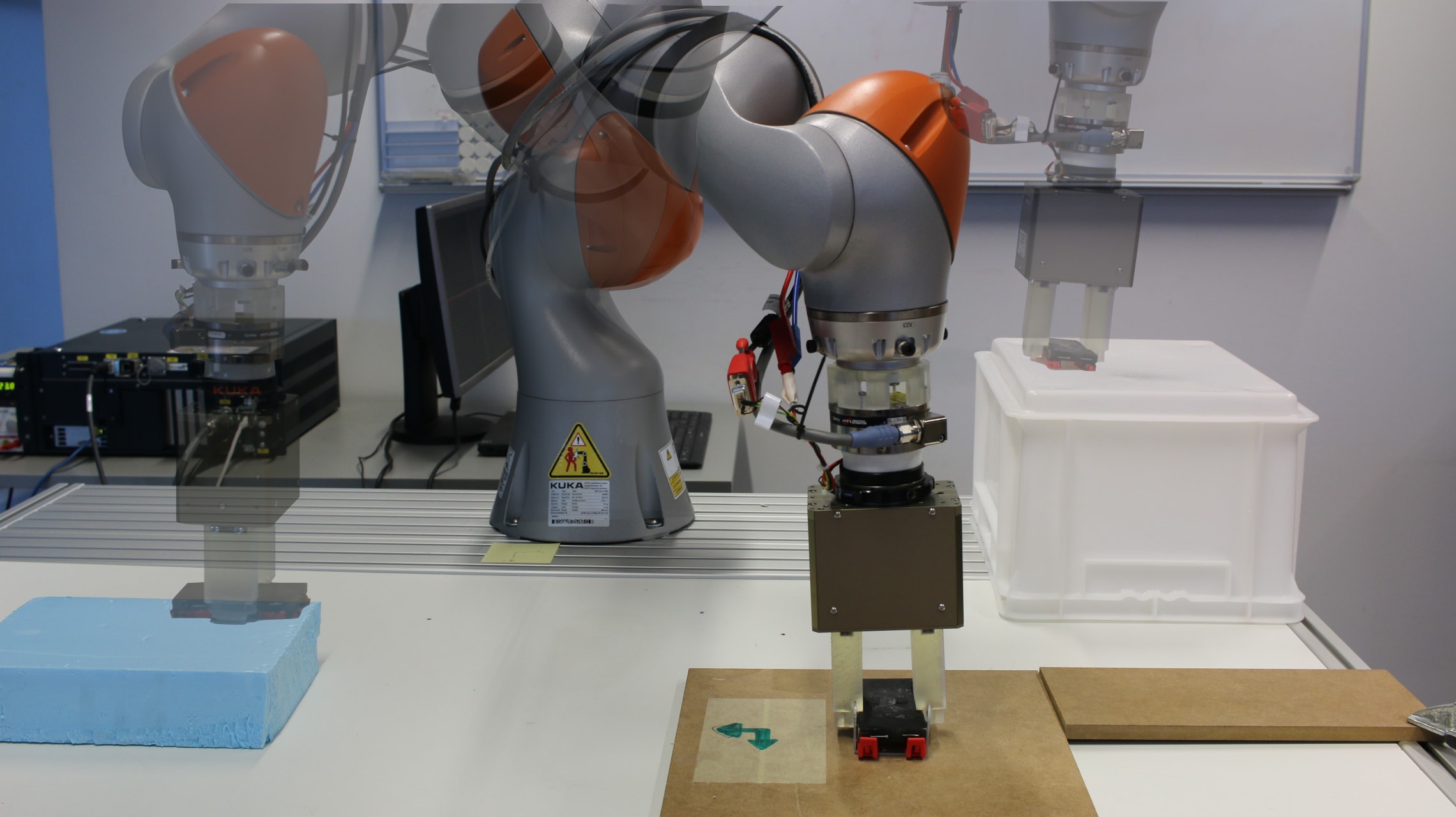}
\caption{The experiment setup. The robot executes learned skill for the task with different initial (shadow robot) 
and goal (solid robot) poses. The data collected are used for training Motion Assessment Primitives (MAPs).}
\label{fig::hardware}
\end{figure}
A KUKA iiwa light-weight robot equipped with an ATI F/T sensor is used for trajectory demonstration and reproduction. We obtain the demonstrated pose via kinesthetic 
teaching, and the corresponding wrench signals from the force/torque sensor mounted at the end-effector. The experiment setup is illustrated in Figure \ref{fig::hardware}. 
Furthermore, Dynamic Movement Primitives (DMPs) \cite{ijspeert2013dynamical} are used to demonstrate
movements which are then reproduced with different initial and goal poses, as illustrated in Fig. \ref{fig::hardware}. 

The proposed method is evaluated on snap-fit assembly and screwing which are common 
force-interaction tasks in manufacturing.  Therefore, we design three toy-cases 
and also use two industrial use-cases of snap-fitting and screwing. The toy-cases include square 
snap-fit assembly (Fig. \ref{fig::square}), round snap-fit assembly (Fig. \ref{fig::round}) and screwing (Fig. \ref{fig::screwing}). The industrial use cases are subtasks of a vehicle's audio-system
assembly provided by Magna Steyr. It consists of a tweeter insertion with snap-fit and speaker screwing which are challenging to perform without force control.

For each toy-case, we collect 3 datasets associated with 3 different demonstrations. 
Each dataset consists of time-stamps, wrench signals, Cartesian poses and labels regarding the outcome of 20 task executions which are used to
train the Naive Bayes classifier. TABLE \ref{tab::over} provides an overview of all the gathered datasets.

The proposed method's generalization ability is evaluated on two 
scenarios: the states' variation and demonstrations' 
variation. In the first case, the classifier is trained from a set of 
movements' generated by DMPs from a specific demonstration. Then the 
classifier 
assesses the outcome of movement generated by the same DMP but with 
different starting and target states. In the more challenging scenario 
of demonstrations' 
variation, the classifier assesses movements which are generated by 
different demonstrations. 

 \begin{table}
 	\renewcommand{\arraystretch}{1.8}
 	\centering
 	 	\caption{Overview of the datasets}
 	\begin{tabular}{|c|c|c|c|c|}
 		\hline
 		\textbf{Task} & \shortstack{ \textbf{Number of}  \\ \textbf{ demos}} & \shortstack{ \textbf{Number of}  \\ \textbf{ trajectories}}  & \shortstack{\textbf{Samples} \\ \textbf{per trajectory}}  \\ \hline
 		Square snap fit & 3 & 60 &97  \\ \hline
 		Round snap-fit   &  3  & 60 & 143 \\ \hline
 		Screwing&  3 & 60 & 204 \\ \hline
 		Twitter snap-fit &  1 & 15 &79 \\ \hline
 		Speaker Screwing &  1 & 15 &178 \\ \hline
 		 
 	\end{tabular}

 	\label{tab::over}
 \end{table}

\subsection{Toy-cases}

\begin{figure}
\centering
\includegraphics[width=0.38\paperwidth]{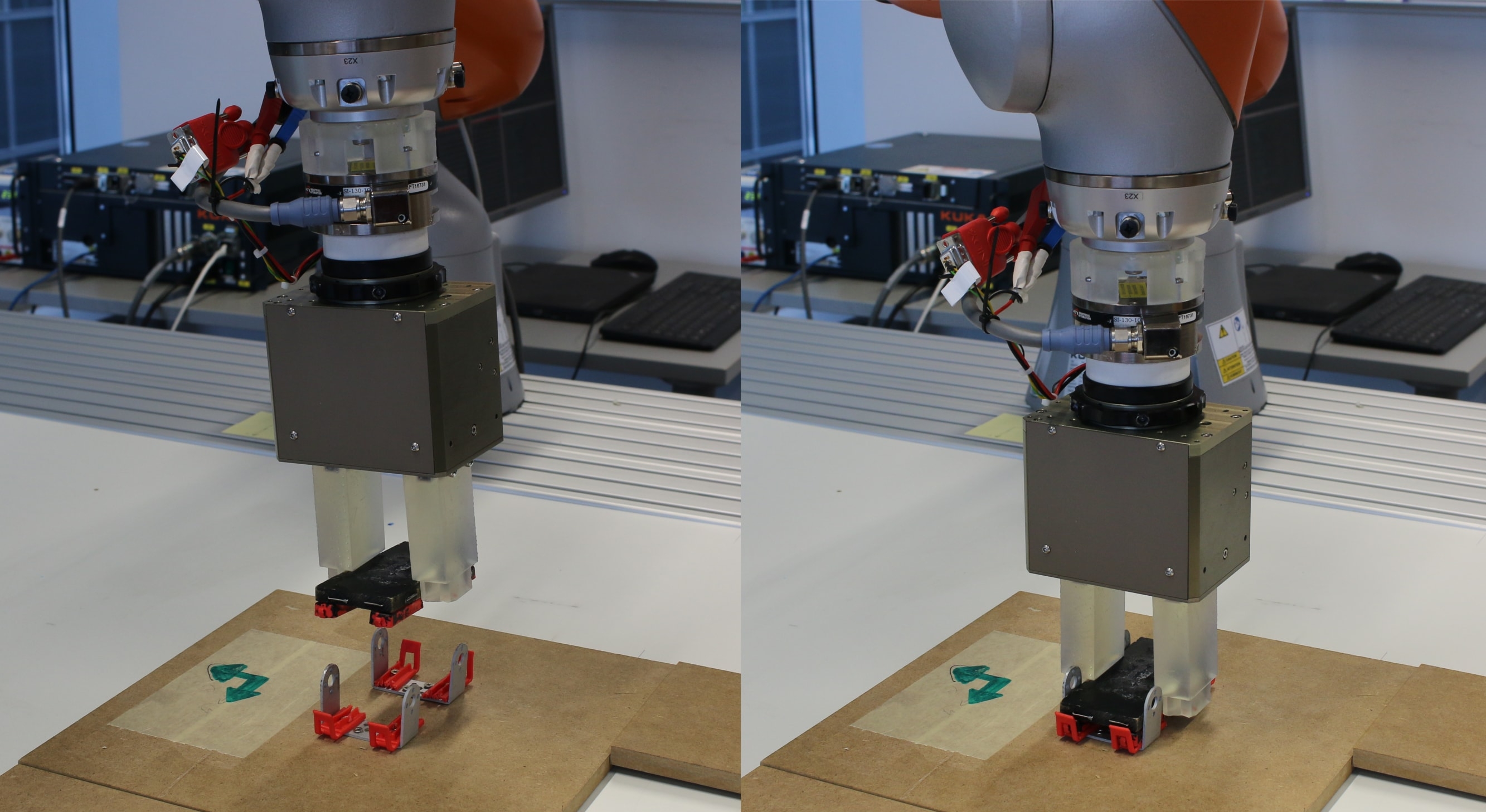}
\caption{Square snap-fit assembly toy-case. The robot inserts the manipulated object into the base with 4 snap-fits (red). The motion is constrained by 4 metal parts (grey) on the base (left). All four snap-fits have to be inserted (right).}
\label{fig::square}
\end{figure}

We design three toy use-cases of varying complexities. For each of those, three demonstrations are performed from different individuals which create three data sets 
to verify the generalization capabilities of MAPs. For each demonstration, 
we first let the robot execute DMPs $ 10 $ times with random initial and goal positions 
that vary on a large scale. We then perform $ 10 $ movement reproductions which result in failure due to inaccurate goal positions or pose variations of the work-pieces. 

The first toy use-case is the square snap-fit assembly. As shown in Fig. \ref{fig::square}, this setup has 4  `L' shaped cantilever snaps (red). 
 On the base, there are  4 metal parts (grey) that constrain the snap-in movement. Thus, this setup only allows 
 up to 3 mm deviation from demonstrated positions during snapping.
 The snap-fits can only be inserted in the  vertical direction with a fixed pose.  
 The  manipulated object (black) to be inserted has notches on its back. If the movement 
 deviates from the demonstration when snapping, the manipulated object  will be jammed 
on the metal parts. When all 4 snap-fits are fully inserted, the execution is successful.

\begin{table}[b]
	\renewcommand{\arraystretch}{1.5}
	\centering
	\caption{Assessment accuracy on all the toy-cases}
	\resizebox{\columnwidth}{!}{
	\begin{tabular}{|c|c|c|c|c|c|}
		\hline
		\textbf{Toy-case:}&\textbf{Generalization to:}&\textbf{Dataset 1 }&\textbf{Dataset 2} &\textbf{Dataset 3}   \\ \hline
		\multirow{2}{*}{Square Snap-fit }
		& Start \& Target States & 90\% & 80\% & 95\% \\ \cline{2-5}
        & Demonstrations         & 75\% & 75\% & 100\%\\ \hline \hline
        \multirow{2}{*}{Round Snap-fit }
        & Start \& Target States & 80\% & 85\% & 75\% \\ \cline{2-5}
        & Demonstrations         & 75\% & 75\% & 70\%\\ \hline \hline
        \multirow{2}{*}{Screwing }
        & Start \& Target States & 80\% & 75\% & 75\% \\ \cline{2-5}
        & Demonstrations         & 85\% & 75\% & 85\%\\ \hline

	\end{tabular}
	}
	
	\label{tab::snap}
\end{table}

 The proposed method's generalization ability to various start/target states and demonstrations for the square snap-fit is included in first two rows of TABLE \ref{tab::snap}.
The movement's generalization to different start/goal states renders very good performance since
dataset 1 and 2 are assessed with more than 90\% of accuracy. As for the generalization to different demonstrations, 
dataset 1 and 2 have good performance with accuracy of more than 75\%. It should be noted that 100\% of success rate is achieved at dataset 3.

\begin{figure}
\centering
\includegraphics[width=0.39\paperwidth]{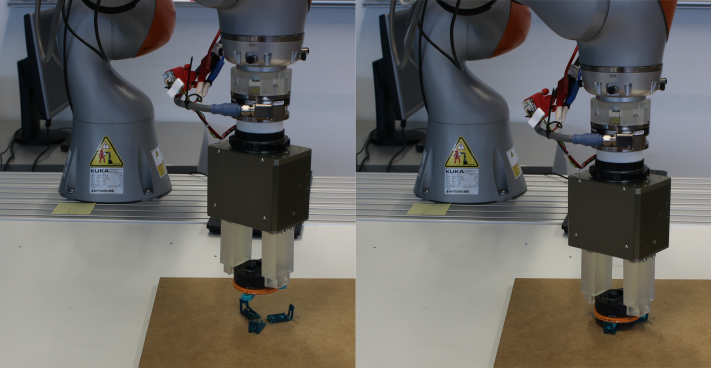}
\caption{The round snap-fit assembly toy-case. The robot inserts the manipulated object (black)
into the base with 3 snap-fits (green). There is no constraints for the motion (left). 
The task is completed when all 3 snap-fits are inserted (right).}
\label{fig::round}
\end{figure}

The second use case is a round snap-fit assembly illustratedin Fig. \ref{fig::round}. 
There are 3 `L' shaped cantilever snaps (green) mounted on a round 
configuration where there is no physical constraint. A challenge of this task is that there are 
different ways of demonstration. In our first demonstration, the user 
firstly inserts one snap and then insert the rest. Likewise, two snaps 
are inserted at first in our second demonstration. Alternatively, all 
three snaps are inserted simultaneously in our third demonstration.
 An execution is classified as a success 
when all snap-fits are inserted.

The proposed method's performance at the case of round snap-fit is illustrated in the middle 
two rows of TABLE \ref{tab::snap}. The start/goal state generalization is more than 80\% for  the 
demonstrations 1 and 2 while it is slightly worse for the third one. Furthermore, the 
generalization to different demonstrations is slightly worse compared to the other tasks.
 This is due to the large variety of demonstrated assembly policies. 

\begin{figure}[b]
\centering
\includegraphics[width=0.39\paperwidth]{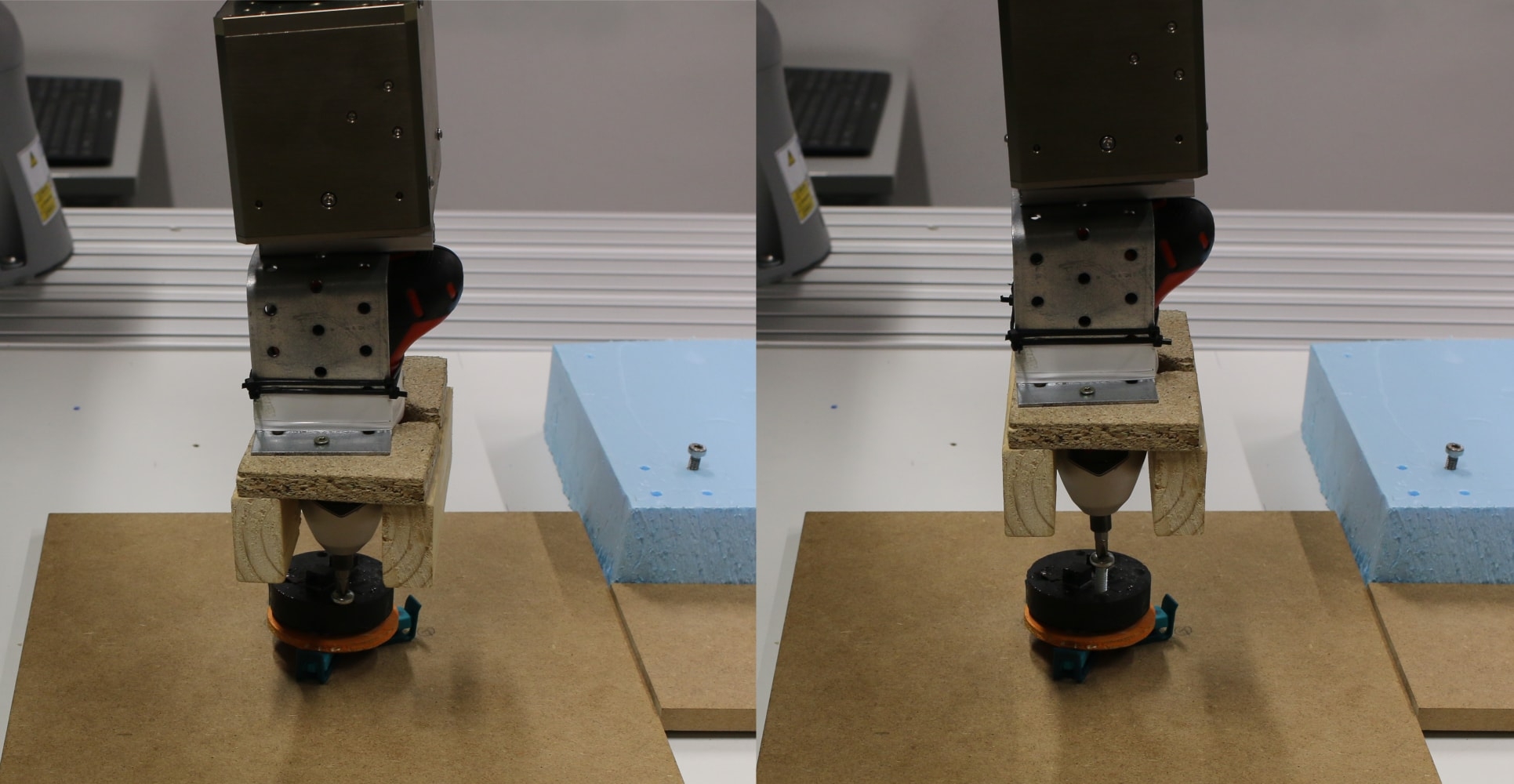}
\caption{The screwing toy-case. The robot picks up a M3.5 screw and screws it at the black base (left).
The Motion Assessment Primitives (MAPs) verify if the screw is tightened in the hole (right).}
\label{fig::screwing}
\end{figure}

The third toy-case is screwing as shown in Fig. \ref{fig::screwing} where a M3.5 bolt is fastened on a 3D printed base (black). The bolt appends on the tip of the 
screw-driver by magnetic force.
In order to activate the screw driver, one must rotate clock-wise and simultaneously apply a vertical force. The bolt
needs to be fully fastened for successful execution.

Screwing is more complex than square snap-fit assembly, since there is no tight 
physical constraint for the movement. Even though the task can only be successfully 
completed in a certain manner, scenarios diverse for failed executions. The tip 
of the screw will have various behaviors if the screw is not inserted in the hole while screwing. 
For instance, the screw may rotate on a flat surface, or the screw may be detached from the tip. Thus failed executions do 
not have similar force profiles which makes this task significantly challenging. Furthermore, since the screw is attached to the tip by magnetic field, 
the pose of the screw is not exactly identical among different movements.
The generalization performance in the screwing toy-case is included in the last two rows of Table \ref{tab::snap}. 

In this challenging task, the generalization to different movement 
states is dropped by about 10\% ranging from 75\% to 80\% compared to 
the square snap-fit. Similarly, the generalization to different 
demonstrations is also reduced.

\subsection{Industrial use-cases}
\begin{figure}[t]
\centering
\includegraphics[width=0.38\paperwidth]{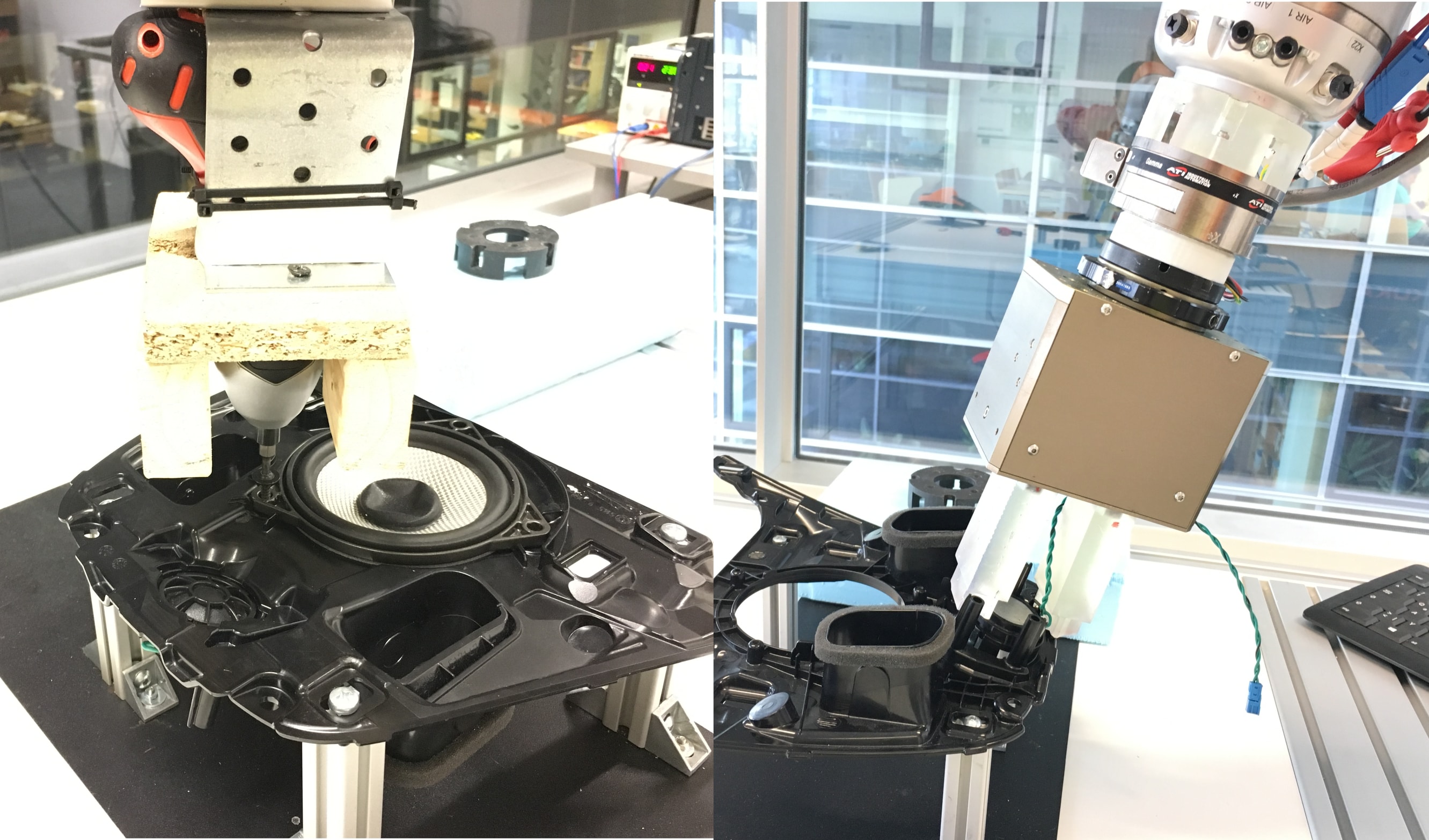}
\caption{The industrial use cases of vehicle audio assembly. A speaker is mounted on a base with 3 screws (left). 
A tweeter is inserted in the base with snap-fits (right).}
\label{fig::Industrial}
\end{figure}

The datasets gathered for toy-cases are created in order to capture a 
large variety of possible failures. Nevertheless, not all of those 
failures take place during normal operating conditions. Therefore, MAPs 
are also evaluated on two industrial use-cases where the robotic system 
operates as it would within a manufacturing scenario. 
The industrial use-cases include screwing and snap-fitting which are common force-interaction tasks in industry. As shown in Fig. \ref{fig::Industrial}, 
we need to insert a tweeter with 3 cantilever snap-fits into the base which has physical constraints 
(right). Concerning screwing, we need to fasten the large speaker on the base with three bolts (right). 

Similar to toy-cases, we perform a demonstration for each task and reproduce it with various start and goal positions. Thus, we perform 15 task executions in order to measure the  prediction accuracy of the MAP on different executed trajectories. The results of the leave-one-out cross-validation are illustrated in Table \ref{tab::industry} where the algorithm's accuracy is significantly increased compared to the toy-cases for both screwing and snap-fittings. This happens because failures during normal operation are much less diverse compared to the failures in the toy-cases datasets.

\begin{table}
\renewcommand{\arraystretch}{1.2}
\centering
\caption{Assessment accuracy on the industrial use cases}
\begin{tabular}{|c|c|c|c|}
\hline
\textbf{Industrial use-case}& \textbf{Success rate}   \\ \hline
Screwing &80\%  \\ \hline
Snap-fitting   &  93.33\%  \\ \hline
\end{tabular}

\label{tab::industry}
\end{table}

Furthermore, from the confusion matrices of the industrial use-cases (Table \ref{tab::truePositive}) 
is observed that, in the case of screwing task, the predictions are not 
biased towards a specific class. Nevertheless, this is not the case for 
snap-fitting due to its unbalanced dataset which contains two failures. 
However, this result demonstrates the proposed method's 
data-efficiency since it is able to correctly assess a failure even with 
a single training instance for a class.

\begin{table}[!htb]
	
\caption{Confusion Matrices for Industrial Use Cases}
\begin{minipage}{.5\linewidth}
	\caption*{Screwing}
	\centering
	\resizebox{\columnwidth}{!}{\renewcommand{\arraystretch}{3}
\begin{tabular}{|c|c|c|c|c|}
	\hline
	$ n=15 $& \shortstack{Predicted: \\ Success } & \shortstack{Predicted: \\ Failure}   \\ \hline
	\shortstack{Actual: \\ Success }& 5 &2 \\ \hline
		\shortstack{Actual: \\ Failure}   &1  &8  \\ \hline
\end{tabular}}
\end{minipage}%
\begin{minipage}{.5\linewidth}
	\centering
	\caption*{Snap-fitting}
		\resizebox{\columnwidth}{!}{\renewcommand{\arraystretch}{3}
\begin{tabular}{|c|c|c|c|c|}
	\hline
	$ n=15 $& \shortstack{Predicted: \\ Success } & \shortstack{Predicted: \\ Failure}   \\ \hline
	\shortstack{Actual: \\ Success } &13 &0 \\ \hline
	\shortstack{Actual: \\ Failure }   &1  &1  \\ \hline
\end{tabular}}
\end{minipage} 
\label{tab::truePositive}
\end{table}



\section{CONCLUSIONS}

In this paper we proposed a novel adaptable, reusable and task-agnostic movement assessment primitive which can be used alongside DMPs in order to form flexible reusable robotic-skills. MAPs assess the 
reproduced task outcome through a Programming by Demonstration (PbD) method similarly to DMPs. The algorithm consists of two learning models, the first compares the similarity between a movement demonstration and reproduction by calculating the statistical distance of their corresponding wrench models which are represented by GPs. The similarity measurements are fed to the second model -- a Naive Bayes classifier -- which assigns the probability of success.

The proposed method has been evaluated on five different use-cases of two diverse force-interaction tasks: screwing and snap-fitting. Furthermore, the generalization ability to various demonstrations and movement reproductions has been assessed on a total of $ 11 $ demonstrations and $ 210 $ trajectories. The experimental results demonstrate an overall good ability of the method to assess motions which are generated either from the same or different demonstrations. Nevertheless, the generalization to various demonstrations appears to be more challenging. Also, the method has to be retrained but not reprogrammed for different tasks (screwing, snap-fitting). This fact significantly increases the re-usability of the proposed primitive which can make it valuable part of skill-based systems. 
Furthermore, MAPs can be used within reinforcement learning algorithms in order to provide a fast estimate for the quality of the performed movement \cite{Polydoros2017}.

Thus, future work can be focused on including such movement assessment methods as evaluation primitives in skill-based systems. Furthermore, the presented method can be expanded with online learning in order to make it capable to assess a predicted motion on-the-fly and also be used as part of a reward function in reinforcement learning.



\addtolength{\textheight}{-12cm}   




\bibliographystyle{ieeetr}
\bibliography{ref}

\end{document}